\documentclass[letterpaper, 10 pt, conference]{ieeeconf}  

\IEEEoverridecommandlockouts                              

\usepackage[acronym]{glossaries}
\usepackage{glossaries-extra}  
\glsdisablehyper
\usepackage{amssymb}
\usepackage{hyperref}            
\usepackage{cleveref}
\usepackage{booktabs}  
\usepackage{caption}
\usepackage{subcaption}
\usepackage{graphicx}
\usepackage{booktabs}
\usepackage{multirow}
\usepackage[T1]{fontenc}
\usepackage[table,xcdraw]{xcolor}
\definecolor{mydarkblue}{rgb}{0,0.08,0.45} 
\usepackage{siunitx}
\renewcommand{\SIrange}[3]{\lbrack\num{#1}\ {;}\ \num{#2}\rbrack\,\si{#3}}
\newcommand{\citep}{\cite}
\newcommand{\citet}{\cite}

\newacronym{nn}{NN}{Neural Network}
\newacronym{rfi}{RFI}{Random Force Injection}
\newacronym{erfi}{ERFI}{Extended Random Force Injection}
\newacronym{rao}{RAO}{Random Actuation Offset}
\newacronym{rl}{RL}{Reinforcement Learning}
\newacronym{dm}{DM}{Domain Randomization}
\newacronym{td3}{TD3}{Twin Delayed Deep Deterministic}
\newacronym{ppo}{PPO}{Proximal Policy Optimization}
\newacronym{gae}{GAE}{Generalized Advantage Estimation}
\newacronym{sea}{SEA}{Series Elastic Actuator}
\newacronym{mlp}{MLP}{Multi-Layer Perceptron}
\newacronym{erfic}{ERFI-C}{Extended Random Force Injection Cumulative}
\newacronym{erfi50}{ERFI-50}{Extended Random Force Injection 50\%}
\newacronym{com}{CoM}{Center of Mass}

\title{\LARGE \bf
Learning and Deploying Robust Locomotion Policies \\  with Minimal Dynamics Randomization
}

\author{
  Luigi Campanaro,
   Siddhant Gangapurwala,
   Wolfgang Merkt, and
   Ioannis Havoutis\\
   Dynamic Robot Systems Group (DRS), University of Oxford\\%
   \texttt{\{luigi,siddhant,wolfgang,ioannis\}@robots.ox.ac.uk}}

\begin{document}

\maketitle
\thispagestyle{empty}
\pagestyle{empty}

\begin{abstract}
Training deep reinforcement learning (DRL) locomotion policies often require massive amounts of data to converge to the desired behavior. In this regard, simulators provide a cheap and abundant source. For successful
sim-to-real transfer, exhaustively engineered approaches 
such as system identification, dynamics 
randomization, and domain adaptation are generally employed. As an alternative,
we investigate a simple strategy of \textit{random force injection} (RFI)
to perturb system dynamics during training. We show that the application of random
forces enables us to emulate dynamics randomization. This allows
us to obtain locomotion policies that are robust to variations in system dynamics.
We further extend RFI, referred to as extended random force injection (ERFI), by introducing an episodic actuation offset. We demonstrate that ERFI provides
additional robustness for variations in system mass offering on average a 53\% improved performance over RFI. 
We also show that ERFI is sufficient to
perform a successful sim-to-real transfer on two different quadrupedal platforms,
ANYmal C and Unitree A1, even for perceptive locomotion over uneven terrain in 
outdoor environments.

\textbf{Additional resources at:} \url{https://sites.google.com/view/erfi-video}

\end{abstract}


\section{Introduction} \label{sec:into}
Deep reinforcement learning (DRL) has emerged as a promising approach for legged robotic control enabling highly dynamic and
sophisticated locomotion capabilities~\citep{Lee2019, yang2020multi, kumar2021rma}. 
The sample complexity associated with high-dimensional problems such as locomotion makes the use of physics simulators~\citep{raisim, makoviychuk2021isaac} appealing for training DRL control policies. 
This convenience, however, often requires addressing the \textit{reality gap} between the simulated training domain and the physical target domain. 


Strategies to address this reality gap often include identification of sensory 
noise which is then modeled and introduced in simulation during 
training~\citep{Jakobi1995, Hwangbo2019}; accurate parameter
identification (of properties such as \glsxtrlong{com} (CoM), mass and inertia of
robot links, impedance gains, system communication delays, and friction) 
for system modeling in addition to identification of
relevant distributions suitable for domain randomization~\citep{Tan2018, Lee2019}; and training a \glsxtrlong{nn} (NN) to model the actuation dynamics of specific actuators, e.g. Series Elastic Actuators \glspl{sea}~\citep{Hwangbo2019, Lee2020}.

\begin{figure}[ht!]
  \centering
  \includegraphics[width=\linewidth]{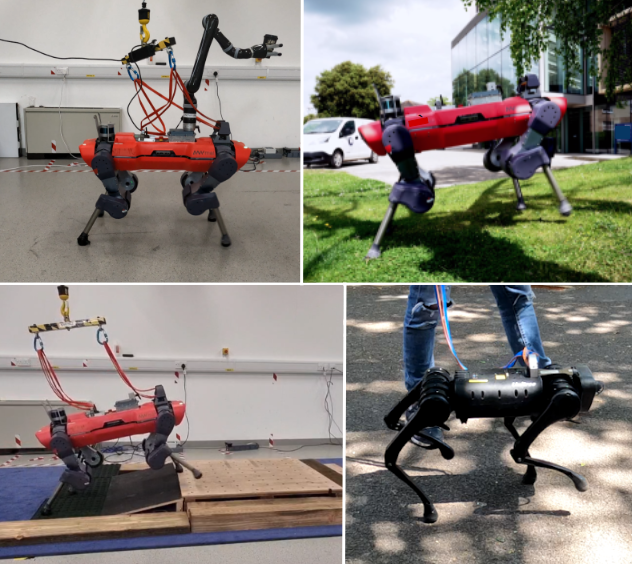}
  \caption{
    Deployment of the perceptive and blind locomotion policies on the ANYmal C and
    Unitree A1 quadrupedal platforms trained using our proposed ERFI-50 strategy
    without requiring actuation modeling or explicit randomization of dynamics or actuation properties.}
   \label{fig:intro_sim_to_real}
   \vspace{-0.4cm}
\end{figure}

As an alternative to exhaustive system identification and distribution identification for dynamics randomization, \citet{Valassakis2020} demonstrated captivating performance in sim-to-real for manipulation tasks using an
extremely simple \gls{rfi} strategy. \gls{rfi} enables emulation of dynamics
randomization through perturbation of system dynamics with randomized forces.
However, as presented in Section~\ref{sec:results},
locomotion policies trained using \gls{rfi} exhibit subpar robustness to policies trained with explicit dynamics randomization. 
To address this loss of performance, we present \gls{erfi}: \gls{erfi} allows to transfer locomotion controllers trained in simulation to the hardware by randomizing only two parameters: a random episodic actuation offset and random perturbations at each step.
First, we show the efficacy of the approach proposed on legged systems, not covered in previous studies, second, we compare it to its predecessor \gls{rfi} \citep{Valassakis2020}, to variations of the same method detailed in the following chapters and to standard domain randomization.
Furthermore, we demonstrate with simulation experiments a significant performance improvement over \gls{rfi} (mass variations' success rate +53\%) especially in unseen scenarios, which involves adding a manipulator arm on top of the robot at test time (mass variations' success rate +61\%).
Finally, we successfully deploy perceptive and blind policies trained in simulation with \gls{erfi} on to the physical ANYmal C and Unitree A1 quadrupeds. We show that training of actuator networks (mainly adopted for robots containing SEAs) and performing significant dynamics randomization~\citep{Peng2018}, currently accepted as a standard for sim-to-real transfer can be substituted by a simple ERFI strategy. We test the controller's locomotion performance over flat and uneven terrain and further evaluate its robustness to additional mass and variation in CoM by mounting a Kinova arm on the robot's base.


\section{Related Works} \label{sec:related_works}
%

Actuators are an essential part of legged systems: they can be hydraulic~\citep{Semini2011}, electric~\citep{6631038} and contain compliant elements~\citep{Hutter2016}. 
Their dynamics is difficult to model involving nonlinear/nonsmooth dissipation, feedback loops and several internal states which are not directly observable.
To accurately approximate \glspl{sea}, the authors of~\citep{Hwangbo2019} trained an actuator network able to output an estimated torque at the joints given as inputs a history of joint position errors and joint velocities recorded from the hardware.
Modeling the actuation dynamics with \glspl{nn}, for robots adopting \glspl{sea}, is now considered a standard and other works employed derivations of the same approach \citep{Gangapurwala2022, Miki2022, Bohez2022}.
The limitations of learning the actuators' dynamics can be summarized in the need of recording motors' torques (not directly measurable for direct drives), training and testing the \gls{nn}. 
In this context it is important to underline that direct drives motors are simpler to model compared to \glspl{sea} and adopting an actuator network is not necessary, since classic system identification is enough. 
However, \gls{erfi} also removes the need for system identification by randomizing motors' torques.

Alongside actuator networks, the rise of highly dynamic controllers is driven by domain randomization, particularly dynamics randomization.
Initially introduced in \citep{Tobin2017, Peng2018}, the approach consists in the randomization of some of the parameters of the robot's dynamics or of the environment.
The additional robustness achieved can then compensate for discrepancies between simulation and the real world.
In~\citep{Hwangbo2019} the domain randomization involves adding noise to the center of mass positions, the masses of links, and joint positions.
In~\citep{Bohez2022} the randomization covers: mass, center of mass position, joint position, joint damping, joint friction, joint position tracking gains $K_p$, torque limits; while regarding the observations' perturbations: delays, joint position noise, angular velocity noise, linear acceleration noise and base orientation noise. 
Alongside with dynamics randomization, observations were also perturbed during training~\citep{Bohez2022} by adding delay, injecting noise into the joint positions, angular velocity, linear acceleration and base orientation.
A significant randomization was also adopted in~\citep{Gangapurwala2022}: gravity, actuation torque scaling, robot link mass scaling, robot link length scaling, random external forces at the base, gravity, actuation torque scaling, link mass scaling, link length scaling, actuation damping gain.

Considering the long list of parameters affected by randomization, the additional robustness it offers requires substantial efforts in system identification: especially in selecting the factors responsible of the reality gap~\citep{Valassakis2020} 
and in defining their randomization range; which if done incorrectly can severely affect the real world performances of the controller, leading to overly conservative policies~\citep{Xie2020}.

An alternative technique -- \glsxtrlong{rfi} -- was proposed in~\citep{Valassakis2020}, it aims to transfer policies trained in simulation to real systems without further tuning, with a limited number of parameters and it consists of injecting random forces into the simulator's dynamics.
This method was tested on manipulation tasks, where it performed comparably to domain randomization.
However, its potential was not evaluated for floating-base systems, especially when the overall stability is compromised by external perturbations.  


\section{Preliminaries}
 
\subsection{System Model}
We model a quadrupedal system as a 
        floating base $B$. The robot state 
        is represented
         w.r.t. a
        reference frame $W$.
        We assume the $z$-axis of $W$, $\mathbf{e}_z^W$,
        aligns with the gravity axis. The base 
        position is then 
        expressed as 
        $r_{B}\in\mathbb{R}^3$, and the 
        orientation, $\mathrm{q}_{B}\in\mathit{SO}(3)$, 
        is represented by a unit quaternion.
        The corresponding rotation matrix is expressed as
        $\mathbf{R}_{B}\in\mathit{SO}(3)$.
        The angular positions of
        the rotational joints in each of the limbs are 
        described by the vector 
        $\mathrm{q}_{j}\in\mathbb{R}^{n_{j}}$.
        For the quadrupeds considered in this work, 
        $n_j=12$. The linear and angular velocities of the base w.r.t. the global
        frame are written as $\mathrm{v}_{B}\in\mathbb{R}^3$ and $\mathrm{\omega}_{\,B}\in\mathbb{R}^3$
        respectively. The generalized coordinates and
        velocities are thus expressed as $\mathrm{q}$ and $\mathrm{u}$ where
        \begin{equation}
            \mathrm{q} = \begin{bmatrix}
                r_{B} \\
                \mathrm{q}_{B} \\
                \mathrm{q}_{j}
            \end{bmatrix} \in\mathit{SE}(3)\times\mathbb{R}^{n_j},\quad \mathrm{u} = \begin{bmatrix}
                \mathrm{v}_{B} \\
                \mathrm{\omega}_{\,B} \\
                \mathrm{\dot{q}}_{j}
            \end{bmatrix} \in\mathbb{R}^{6 + n_j}.
        \end{equation}

\subsection{Impedance Control}
    In the context of this work, we consider
    a quadrupedal 
    system is actuated using the joint 
    control torques $\tau_j\in\mathbb{R}^{n_j}$.
    These torques are computed using the
    impedance control model given by
    \begin{equation}
        \tau_j = K_p(\mathrm{q}^{\ast}_j-\mathrm{q}_j) + 
        K_d(\mathrm{\dot{q}}^{\ast}_j-\mathrm{\dot{q}}_j) 
        + \tau_{j_{FF}},
    \label{eq:impedance_controller}
    \end{equation}
    where $K_p$ and $K_d$ refer to the position and 
    velocity tracking gains respectively, 
    $\mathrm{q}^{\ast}_j$ is the
    vector representing desired joint positions, 
    $\mathrm{\dot{q}}^{\ast}_j$, the desired joint 
    velocities, and $\tau_{j_{FF}}$ refers to the
    feed-forward joint torques. 
    
    For locomotion, we train DRL control policies that modulate
    the joint actuation torques by generating $\mathrm{q}^{\ast}_j$.
    Additionally, we set $\mathrm{\dot{q}}^{\ast}_j=0$ and 
    $\tau_{j_{FF}}=0$. \citet{peng2017learning} presented that such
    an approach offers more stable training and better performance than a torque
    controller. Equation~\ref{eq:impedance_controller} can thus
    be simplified to
    \begin{equation}
        \tau_j = K_p(\mathrm{q}^{\ast}_j-\mathrm{q}_j)
        - K_d\mathrm{\dot{q}}_j.
    \label{eq:impedance_controller_simplified}
    \end{equation}
\subsection{Rigid Body Dynamics Model}
The rigid body dynamics model of a quadrupedal system can be expressed in the form of
generalized equations of motion expressed as
\begin{equation}
        \mathrm{M}\dot{\mathrm{u}} + \mathrm{h} = \mathbf{S}^T\tau_j + 
        \mathrm{J}^{T}\lambda,
        \label{equation:methodology:rigid_body_dynamics:full}
    \end{equation}
    where $\mathrm{M}\in\mathbb{R}^{(6+{n_j})\times(6+{n_j})}$ is the mass matrix relative to the joints, 
    $\mathrm{h}\in\mathbb{R}^{6+{n_j}}$ comprises Coriolis, centrifugal and gravity terms,
    $\mathbf{S}^T = [\mathbf{0}_{{n_j} \times 6} \enspace \mathbf{I}_{{n_j}\times{n_j}}]^T$, and $\mathrm{J}$ is the Jacobian
    which maps the contact forces $\lambda\in\mathbb{R}^{n_f}$ at ${n_f}=4$ feet to generalized forces.


\section{Extended Random Force Injection} \label{sec:extended_random_force_injection}
\citet{Valassakis2020} investigated the effects of introducing random perturbations to a manipulation
system. These \textit{random force injections} aimed to diversify the visited states during training 
of DRL policies. In this regard, their implementation augmented the generalized equations of motion,
similar to Equation~\ref{equation:methodology:rigid_body_dynamics:full},
by random forces $f_r\sim\mathcal{U}(-f^{lim}_r, f^{lim}_r)$ sampled from a uniform distribution $\mathcal{U}$
with limits $-f^{lim}_r$ and $f^{lim}_r$. These forces are
sampled and applied at each time step to perturb the state transition $P$. 
In this work, we adapt this approach for quadrupedal systems and write 
Equation~\ref{equation:methodology:rigid_body_dynamics:full} with RFI as 
\begin{equation}
        \mathrm{M}\dot{\mathrm{u}} + \mathrm{h} = \mathbf{S}^T\tau_j + 
        \mathrm{J}^{T}\lambda + f_r.
        \label{equation:methodology:rigid_body_dynamics:rfi_original}
    \end{equation}
It is important to note that \citet{Valassakis2020} used this approach for a fixed-base system. In our preliminary
experiments, for a mobile-base quadrupedal system, we observed that perturbing the robot's base with even small
forces and torques resulted in convergence to undesired locomotion behavior. For the ANYmal C quadruped, 
forces and torques on the base sampled from distributions with
$f^{lim}_{r_b}>\SI{5}{\newton}$ and $\tau^{lim}_{r_b}>\SI{3}{\newton\metre}$ respectively resulted in
pronking behavior. Although this behavior was robust to external disturbances on the base, 
the pronking gait is energy inefficient and unsuitable for transfer 
to the physical system. Therefore, to better handle uncertainty in the system, we only introduce perturbations to the rotary joints of the quadruped and randomize forces on DoFs that we directly control.

The impedance controller described by Equation~\ref{eq:impedance_controller} is often executed 
at the actuation level at a higher frequency compared to the locomotion controller which is
described by the DRL control policy mapping robot state information to desired joint positions.
In this article, we refer to these frequencies as impedance control frequency and locomotion control
frequency. We introduce perturbations at the impedance control frequency. We then split
Equation~\ref{equation:methodology:rigid_body_dynamics:rfi_original},
describing \textbf{RFI}, into the generalized
equations of motion given by Equation~\ref{equation:methodology:rigid_body_dynamics:full} and an
augmented impedance controller given by
    \begin{equation}
        \tau_j^r = K_p(\mathrm{q}^{\ast}_j-\mathrm{q}_j)
        - K_d\mathrm{\dot{q}}_j + \tau_{r_j},
    \label{eq:impedance_controller_simplified_rfi}
    \end{equation}
where $\tau_{r_j}$ refers to the random joint torque injections sampled
from $\mathcal{U}(-\tau^{lim}_{r_j}, \tau^{lim}_{r_j})$ at each impedance  control update step. Note that
Equation~\ref{eq:impedance_controller_simplified_rfi} is only utilized during training.
For deployment, we consider the actuation is governed
by Equation~\ref{eq:impedance_controller_simplified}.

In this work, we also investigate the effects of introduction of episodic actuation offsets during training.
As opposed to randomizing $\tau_{r_j}$ at each impedance control step, we sample joint torque offsets  $\tau_{o_j}$ from $\mathcal{U}(-\tau^{lim}_{o_j}, \tau^{lim}_{o_j})$, 
at the beginning of each training episode and apply them at each impedance control step. We refer to this as random actuation offset (\textbf{RAO}). This can be represented similarly as our implementation of \gls{rfi} and
is written as
    \begin{equation}
        \tau_j^o = K_p(\mathrm{q}^{\ast}_j-\mathrm{q}_j)
        - K_d\mathrm{\dot{q}}_j + \tau_{o_j}.
    \label{eq:impedance_controller_simplified_rao}
    \end{equation}
This constant offset enables us to emulate a shift in the robot's mass, inertia, impedance gains
and contact Jacobian. However, unlike \gls{rfi}, wherein the dynamics vary at each impedance control 
step resulting in a more reactive control behavior robust
to temporally local perturbations, with \gls{rao}, the policy learns an implicit
adaptive behavior for temporally global variations in system dynamics.

We also introduce an extended variant of \gls{rfi} by combining \gls{rfi} and \gls{rao} to learn control policies 
which can be robust to temporally local and global variations in system dynamics.
We refer to this as \textbf{ERFI-C}. In this case,
we inject both a randomized force sampled at each impedance control step and an episodic actuation offset. The impedance
controller with \gls{erfic} can be then written as
    \begin{equation}
        \tau_j^c = K_p(\mathrm{q}^{\ast}_j-\mathrm{q}_j)
        - K_d\mathrm{\dot{q}}_j + \tau_{r_j} + \tau_{o_j}.
    \label{eq:impedance_controller_simplified_erfic}
    \end{equation}
We further explore another strategy with the same motivation as for \gls{erfic}. In this case, we only utilize
RFI with 50\% of the parallelized DRL training environments. 
The remaining environments employ \gls{rao}. We refer to this approach as \textbf{ERFI-50}. In comparison to \gls{erfic}, which can be considered as 
\gls{rfi} with randomized distribution mean thereby resulting in a possibility
of a learning bias for robustness to temporally local perturbations, 
\gls{erfi50} promotes unbiased learning of both local and global
variations in system dynamics.

\begin{figure*} [ht!]
    \begin{subfigure}{0.48\textwidth}
      \includegraphics[width=\linewidth]{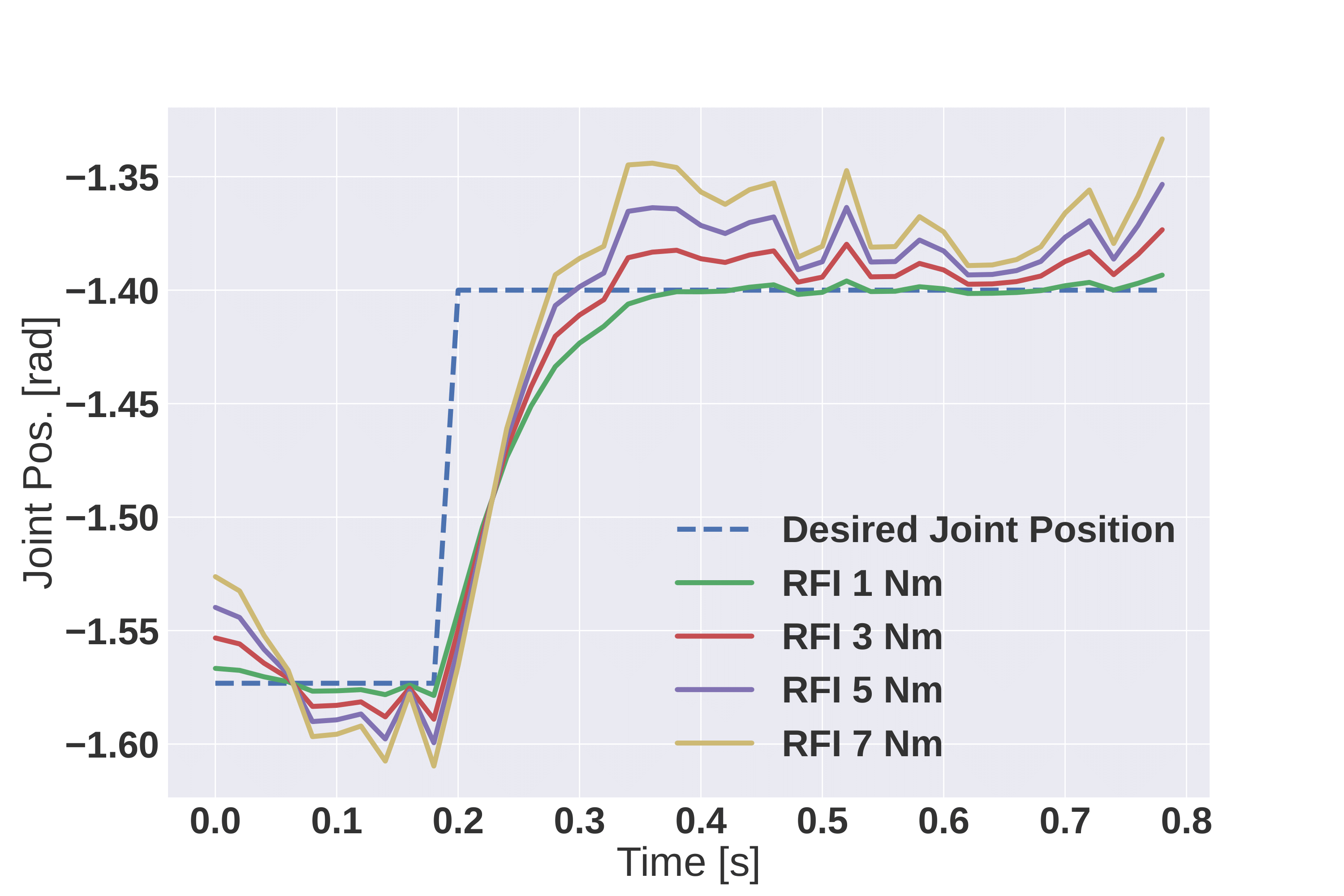}
      \caption{}
      \label{fig:understanding_RFI}
    \end{subfigure}
    \hfill
    \begin{subfigure}{0.48\textwidth}
      \includegraphics[width=\linewidth]{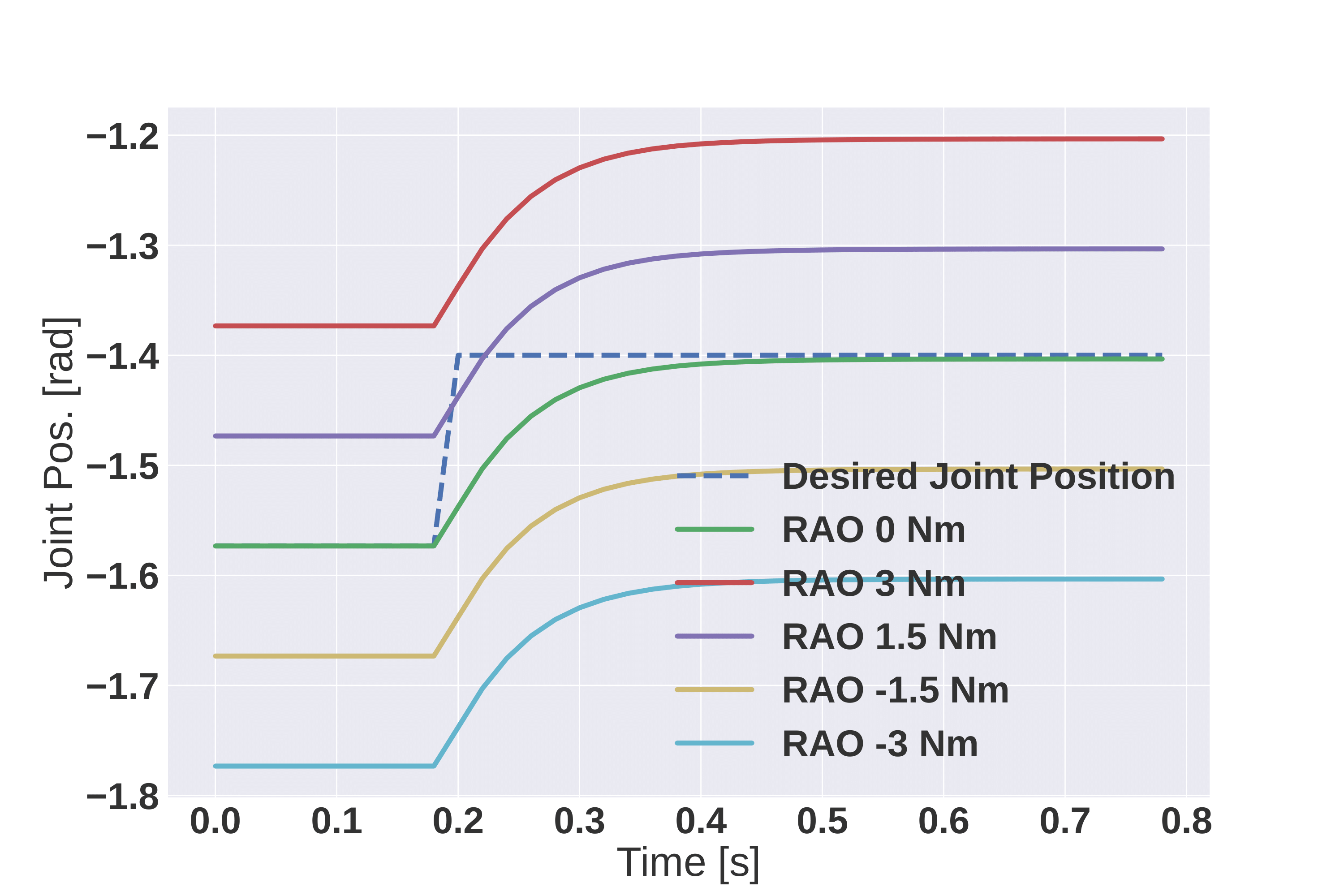}
      \caption{}
      \label{fig:understanding_RAO}
    \end{subfigure}
    \caption{The magnitudes of $\tau^{lim}_{r_j}$ and $\tau^{lim}_{o_j}$ affect the dynamics of the system.}
    \label{fig:understanding_ERFI} 
    \vspace{-0.1cm}
\end{figure*}

\section{Why does ERFI work?}
In \Cref{fig:understanding_RFI} and \Cref{fig:understanding_RAO}, respectively, we show the effects of adding \gls{rfi} and \gls{rao}  as a feed-forward term of the PD controller ($K_p=15$, $K_d=1$) when commanding a step position change of \SI{+0.17}{\radian} (\SI{\approx10}{\deg}) to the hind right knee. 

\subsection{How does RFI model delays?}
As can bee seen from \Cref{fig:understanding_RFI}, the yellow line reaches the desired position faster than the green line, although the green line settles earlier. This implies that RFI adds stochasticity to the rise and settling times, i.e. it either increases or reduces the rise and settling times. The increase or decrease depends on the direction of the perturbation. This allows us to implicitly randomise actuation dynamics, especially parameters that relate to delays, friction and inertia (Section "ERFI robustness to delays" of the accompanying website).

\subsection{How does RAO model mass and kinematic variations?}
In \Cref{fig:understanding_RAO}, the additional torque shifts the desired position of the joint and implicitly models offsets in the joint position (kinematics variations) or in the payload supported by the robot. Evidences of these effects can be found in \Cref{fig:mass_success_rate_no_arm} and \Cref{fig:mass_success_rate_arm} and video 3, 4, and 10 (on the accompanying website), demonstrating the robustness of the controllers even when the unmodelled payload reaches 42\% of the total weight of the robot.


\section{Problem Definition}
The complex \glspl{sea} present on the ANYmal C quadruped exhibit a highly nonlinear behavior \citep{Gehring2016}. To address their complex dynamics, networks modeling the actuation became common practice in the community (\Cref{sec:related_works}). 
Evaluating the effectiveness of \gls{erfi} on such a platform thus provides a measure of the robustness of the method and of its generalization abilities.

Conversely, Unitree's A1 adopts quasi-direct drive actuators, which are affected by high levels of delay, signal noise and inaccurate tracking.
Given the different technology adopted compared to \glspl{sea} and the canonical role that Unitree A1 has played in recent research works \citep{Shao2022, Yang2022}, we also investigated the effects of \gls{erfi} for obtaining locomotion policies for A1.

\subsection{Perceptive Quadrupedal Locomotion} \label{sec:perceptive_locomotion}

The ANYmal C robot is used to track a velocity command ${[v_x, v_y, \dot{\gamma}]}_\mathcal{B}$ on uneven ground using proprioceptive and exteroceptive information.
The state is represented as $s := \langle s_r, s_v, s_{j_p}, s_{j_v}, s_a, s_m, s_c \rangle$, where $s \in {\mathbb{R}}^{259}$, $s_r^\mathcal{B} \in {\mathbb{R}}^{3}$ is the second row of the rotation matrix, $s_v \in {\mathbb{R}}^{6}$ is the base linear and angular velocities, $s_{j_p}^\mathcal{B} \in {\mathbb{R}}^{24}$ is the sparse history of joint position errors and $s_{j_v}^\mathcal{B} \in {\mathbb{R}}^{24}$ is the sparse history of joint velocities, $s_a \in {\mathbb{R}}^{12}$ is the previous action, $s_m \in {\mathbb{R}}^{187}$ are measurements from the height-map around the robot's base and $s_c^\mathcal{B} \in {\mathbb{R}}^{3}$ is the velocity command. 
The actions $a \in {\mathbb{R}}^{12}$ are interpreted as the reference joint positions $\mathrm{q}^*_j$. 
The state $s$ is fed to an \gls{mlp} network made by three layers respectively of size $ [512, 256, 128] $ and the action $a$ is subsequently tracked by the low level PD controller ($K_p = 80., K_d = 2.$).

\subsection{Blind Quadrupedal Locomotion}
The A1 quadruped robot is required to follow a velocity command ${[v_x, v_y, \dot{\gamma}]}_\mathcal{B}$ on flat ground using proprioceptive information.
The state is represented as $s := \langle s_r, s_v,  s_{j_p}, s_{j_v}, s_a, s_c \rangle$, where $s \in {\mathbb{R}}^{192}$, $s_r^\mathcal{B} \in {\mathbb{R}}^{3}$ is the second row of the rotation matrix, $s_v \in {\mathbb{R}}^{6}$ is the base linear and angular velocities, 
$s_{j_p}^\mathcal{B} \in {\mathbb{R}}^{84}$ is the history of joint position errors and $s_{j_v}^\mathcal{B} \in {\mathbb{R}}^{84}$ is the history of joint velocities, $s_a \in {\mathbb{R}}^{12}$ is the previous action, velocity and action and $s_c^\mathcal{B} \in {\mathbb{R}}^{3}$ is the velocity command. 
The actions $a \in {\mathbb{R}}^{12}$ are interpreted as the reference joint positions $\mathrm{q}^*_j$. 
The state $s$ is fed to an \gls{mlp} network formed by two layers respectively of size $ [512, 512] $ and the action $a$ is tracked by the low level PD controller ($K_p = 15., K_d = 1.$).
The base linear velocity in $s_v$ is not provided by the onboard state estimator and it was estimated similarly to \citep{Ji2022} through an \gls{mlp} network of size $[128, 128]$.

\begin{figure*}[ht!]
  \centering
  \includegraphics[width=0.9\linewidth]{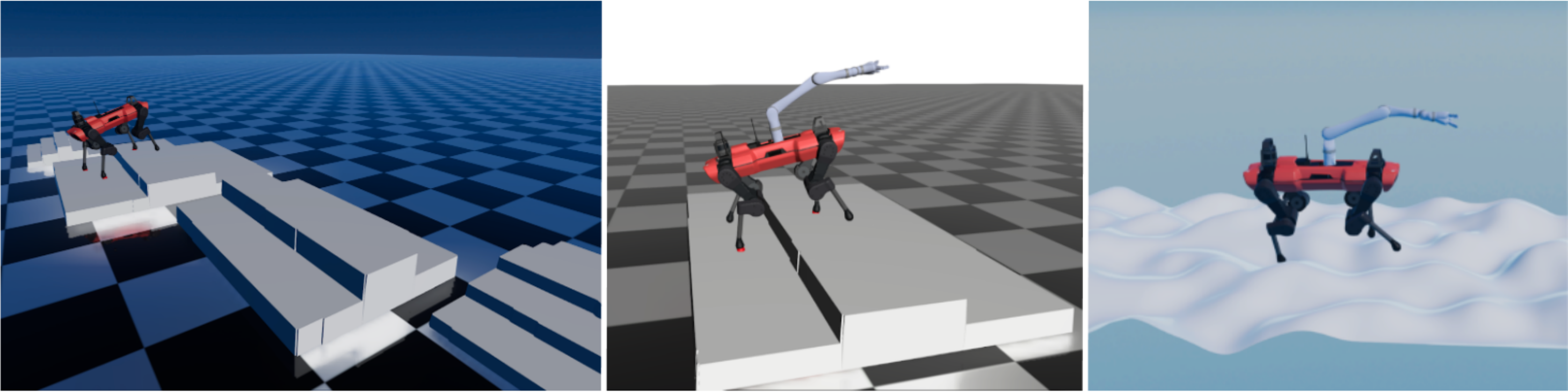}
  \caption{(Left) Examples of stairs with varying step-height and step-depth used for 
  evaluation. (Center) ANYmal C walking on stairs with an unmodeled Kinova 
  manipulator. (Right) ANYmal C walking on rocky terrain during tests.}
   \label{fig:test_environments}
  \vspace{-0.3cm}
\end{figure*}

\section{Experimental Setup} \label{sec:experimental_design}
To evaluate our method, we employed ANYmal C as a reference platform and we trained different policies for 10,000 iterations using IsaacGym~\citep{Makoviychuk2021} each adopting one among \gls{rfi}, \gls{rao}, \gls{erfi50}, \gls{erfic}, and ActNetRand, where ActNetRand represents the present state-of-the-art approach implementing both actuation network and extensive domain randomization, as in \citep{Rudin2021}.
The environment settings used are described in \Cref{sec:perceptive_locomotion}. 

The performances of the policies trained with the different methods were assessed by addressing perceptive locomotion over stairs and rocky terrain as relevant case study (\Cref{fig:test_environments}). 
Moreover, to obtain more realistic results the experiments were conducted in a different simulator (RaiSim~\citep{raisim2018}) and we included an actuator network to reproduce the dynamics of \glspl{sea} (which was not used during the training of RFI/RAO/EFRI policies).
The robot is always deployed at the same position, the velocity command is fixed to $\SI{0.5}{\meter\per\second}$ and it has $\SI{8}{\second}$ to go up the stairs; the attempt is considered a failure when the robot falls on the ground or when it is not able to move forward for at least \SI{2.5}{\meter}.
We generated 50 random stairs as in~\citep{Gangapurwala2022}, which are placed just in front of the robot, and walking for \SI{2.5}{\meter} from the spawning point requires tackling at least one step.

To assess the robustness of the policies to unseen conditions we introduced perturbations to the simulation environment: the application of external forces \SIrange{0}{150}{\newton} to the base (fixed value during training: 0.) for a duration of \SI{3}{\second}, the application time of an external force of \SI{50}{\newton} varies between \SIrange{0}{3}{\second} (fixed value during training: 0.), the application of external torque \SIrange{0}{75}{\newton\metre} to the base (fixed value during training: 0.) for a duration of \SI{1}{\second}, the friction coefficient between ground and feet in the range $[0.2, 0.8]$ (fixed value during training: 0.5), the gravitational acceleration was modified between \SIrange{-18}{-2}{\meter\per\second^2} (fixed value during training:  \SI{-9.81}{\meter\per\second^2}), the position of the knees' motors was shifted by \SIrange{-0.15}{-0.15}{\meter} (fixed value during training: 0.) and the mass of the base changed between \SIrange{22}{65}{\kilogram} (fixed value during training: \SI{27}{\kilogram}).
Throughout the evaluation we alter only one parameter at the time and for each of them we run 50 experiments with different terrains.
Furthermore, we replicate the set of experiments above with a robotic arm mounted on top of ANYmal C, this introduces significant variations in the mass matrix $M$ which the robot never explicitly experienced during training.
Following the thorough validation presented in simulation, the best performing controller (resulted to be trained with \gls{erfi50}) was deployed on the hardware, tested on rough and uneven terrain, both in the laboratory and outdoor environments to validate the feasibility of the method.

\begin{figure*}[ht!]
  \centering
  \includegraphics[width=0.95\linewidth]{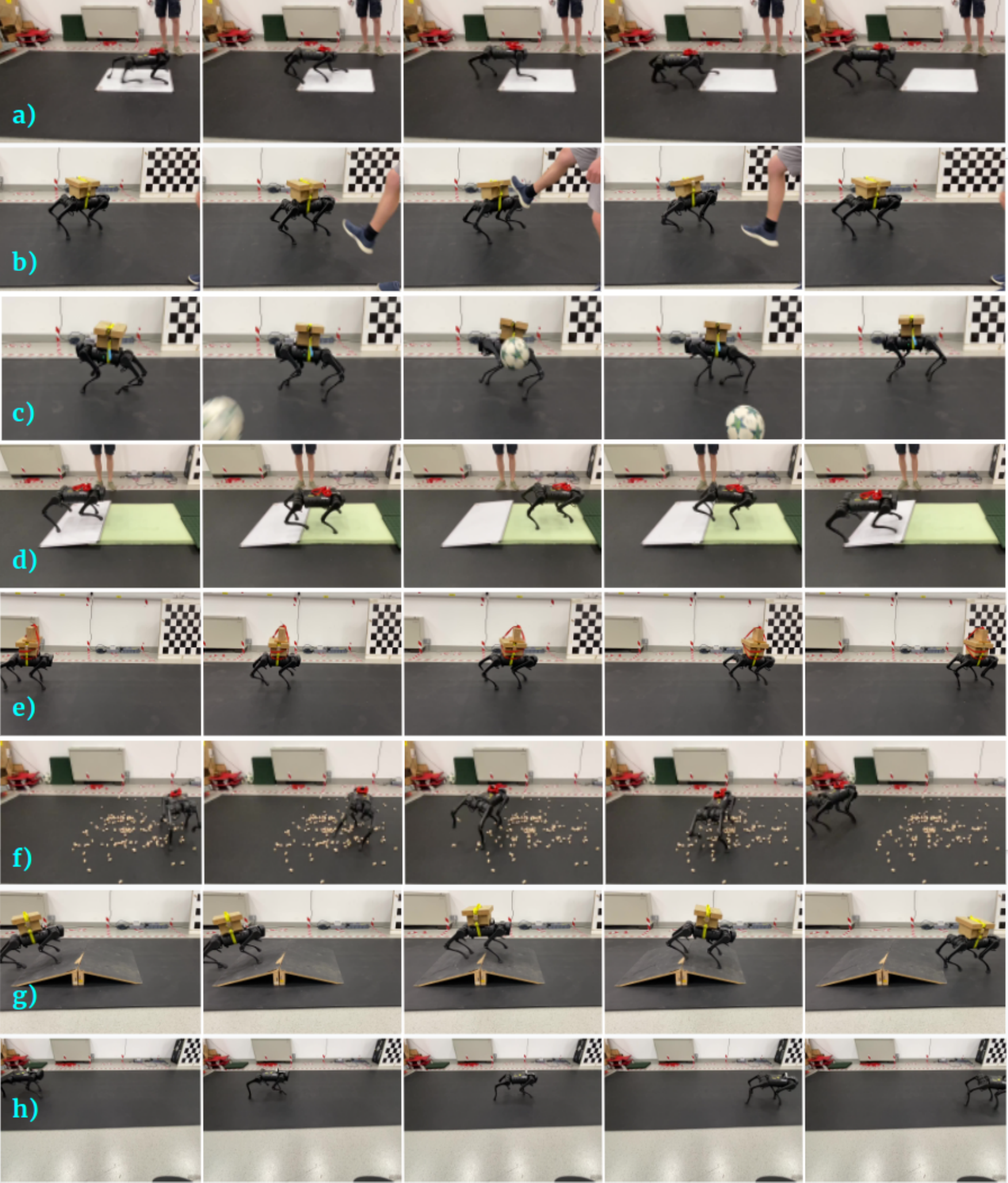}
  \caption{This figure shows some of the experiments on Unitree A1 adopting ERFI, also part of our accompanying website \url{https://sites.google.com/view/erfi-video}. a) Walking on wet terrain and recovering from slipping, b) resisting to external forces, c) withstanding impulsive forces, d) walking on soft terrain, e) walking with an unknown 5 Kg payload, f) walking on wooden cylinders, g) traversing a ramp, and h) adapting to a $K_p^{RH KFE}$ equal to a third of the original value.}
   \label{fig:a1_experiments}
\end{figure*}

In addition, we demonstrate the effectiveness of \gls{erfi50} with hardware experiments also on Unitree A1, this time performing blind locomotion in challenging conditions.
We present results of extensive hardware evaluation in \Cref{fig:a1_experiments} and on  our accompanying website \url{https://sites.google.com/view/erfi-video}.

\begin{figure*} [ht!]
    \begin{subfigure}{.32\textwidth}
      \centering
      \includegraphics[width=.95\linewidth]{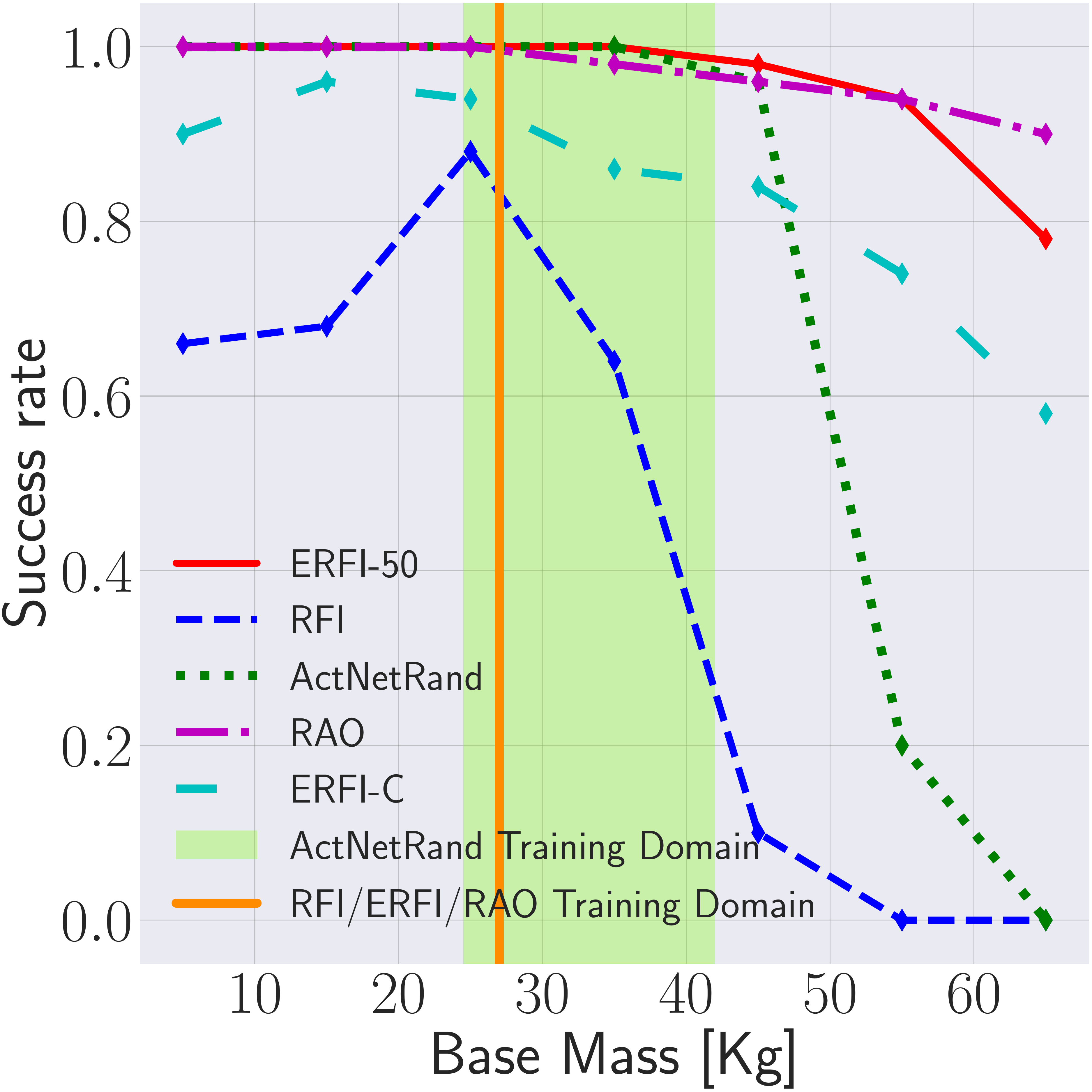}
      \caption{}
      \label{fig:mass_success_rate_no_arm}
    \end{subfigure}
    \begin{subfigure}{.32\textwidth}
      \centering
      \includegraphics[width=.95\linewidth]{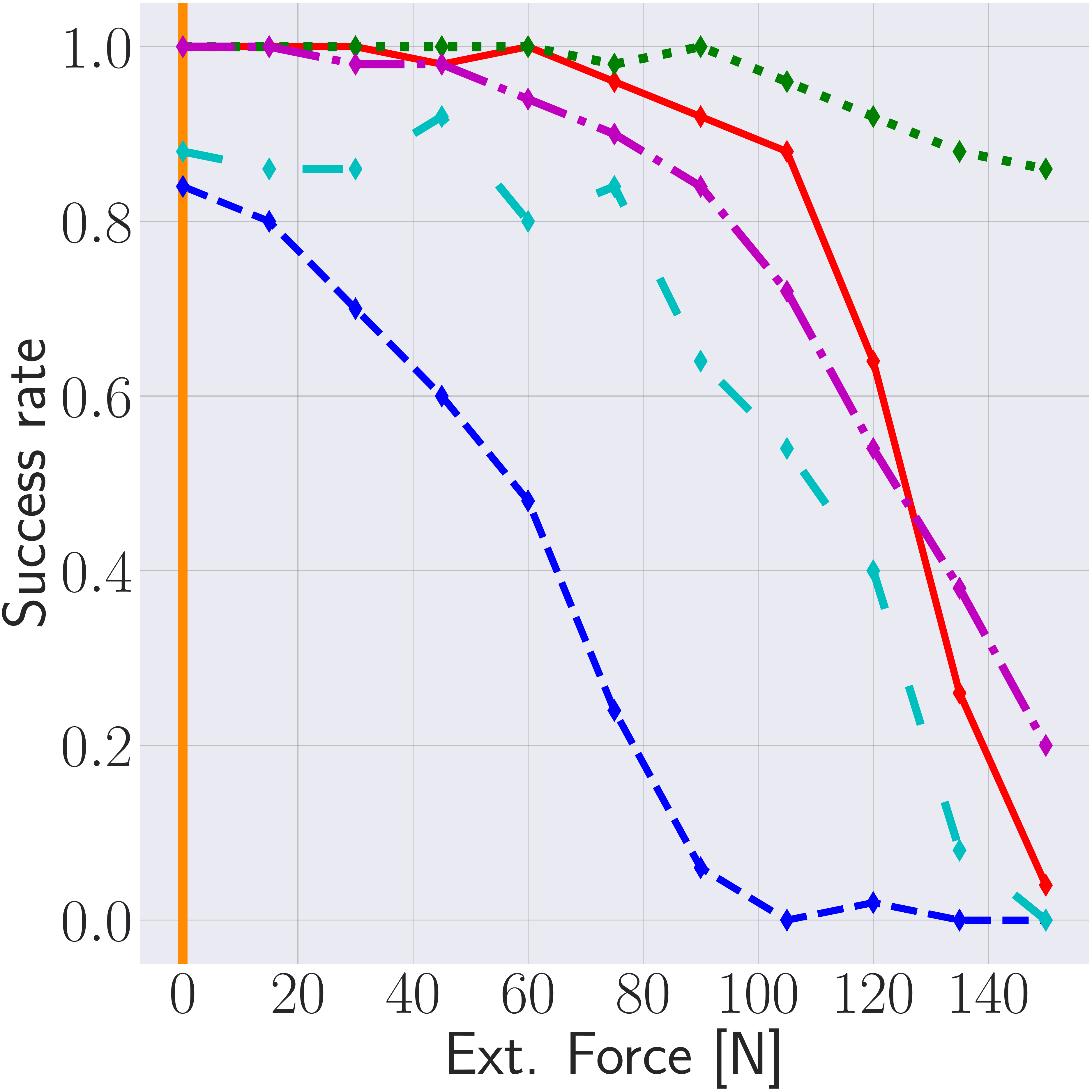}
      \caption{}
      \label{fig:extForceMagnitude_success_rate_no_arm}
    \end{subfigure}
    \begin{subfigure}{.32\textwidth}
      \centering
      \includegraphics[width=.95\linewidth]{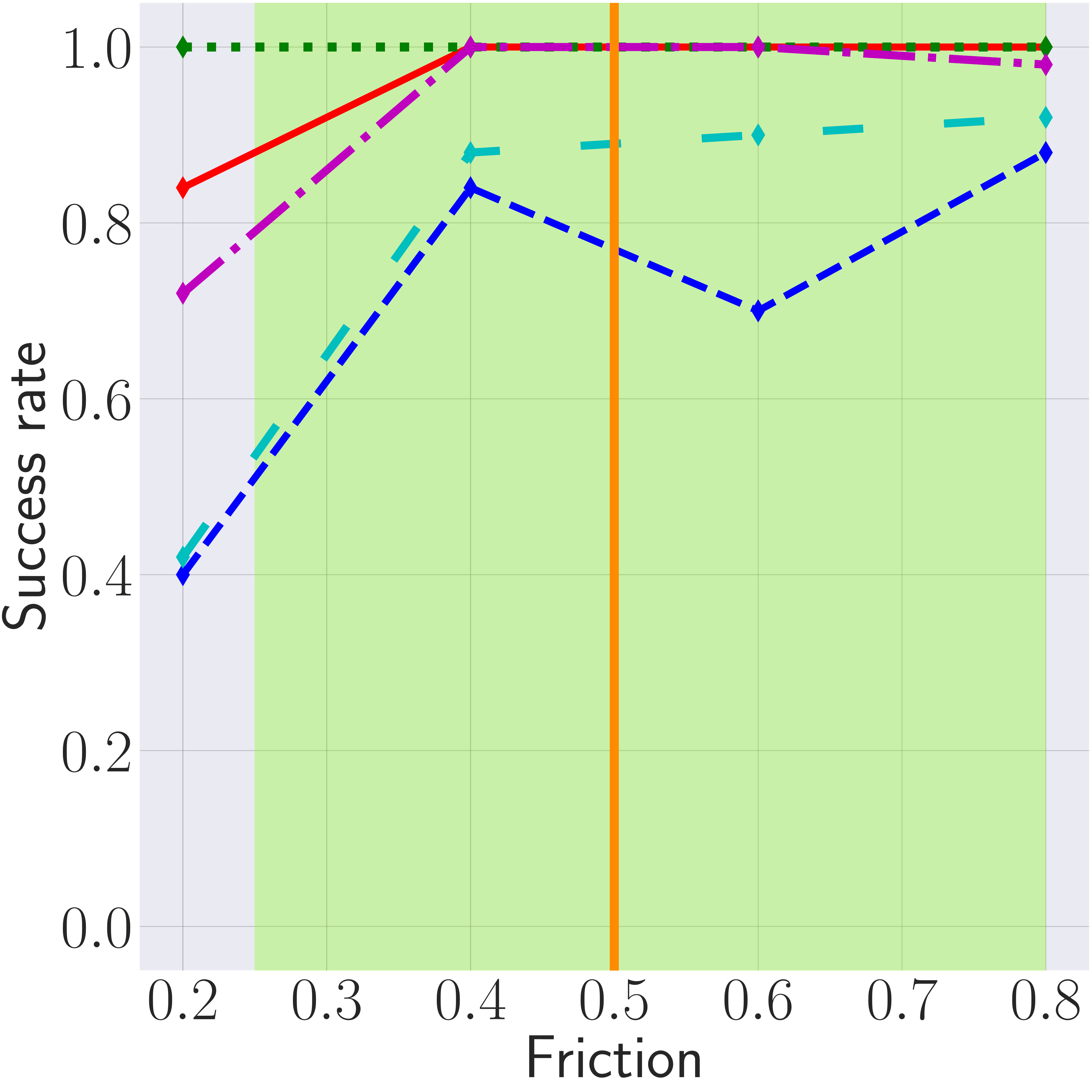}
      \caption{}
      \label{fig:friction_success_rate_no_arm}
    \end{subfigure}
    \\
    \begin{subfigure}{.32\textwidth}
      \centering
      \includegraphics[width=.95\linewidth]{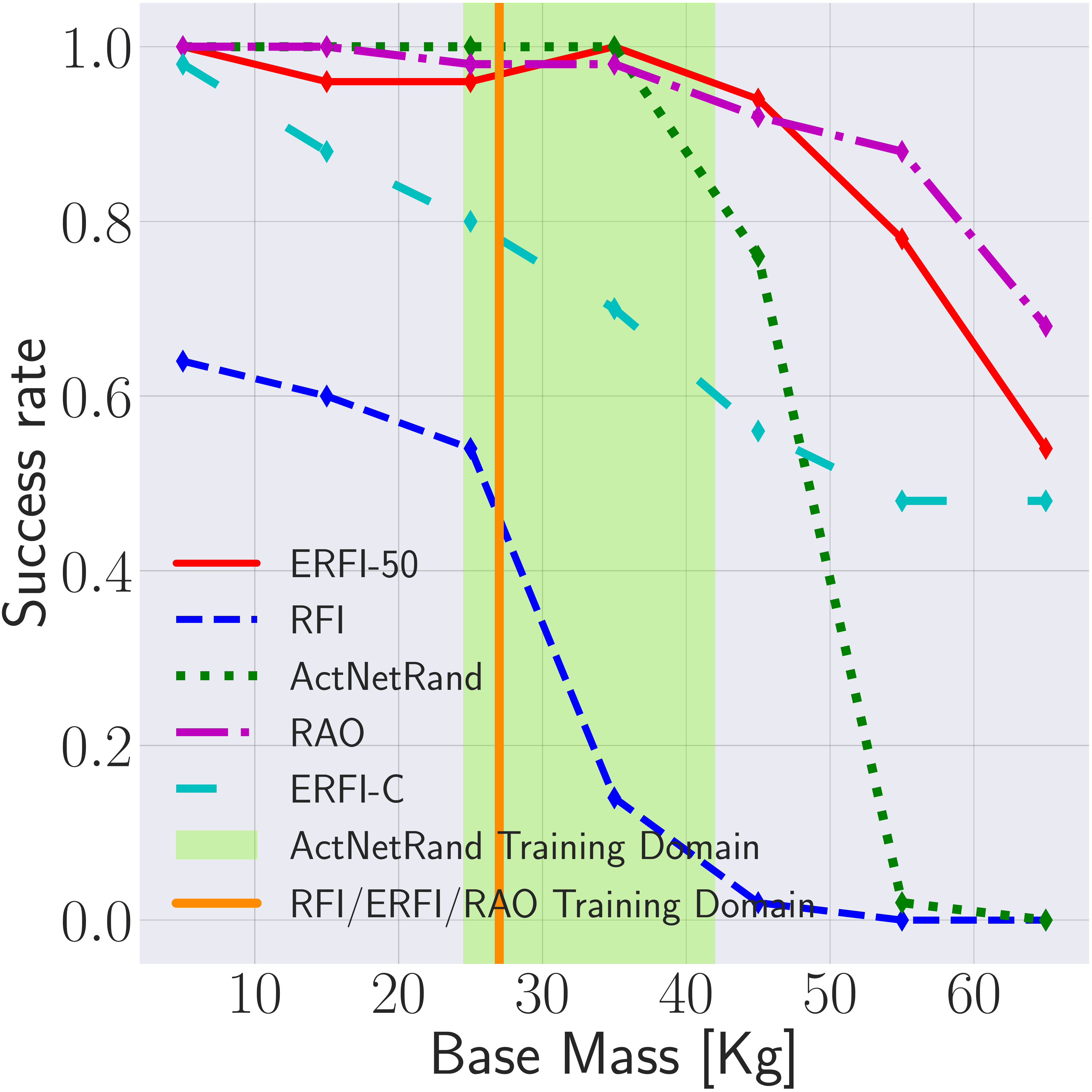}
      \caption{}
      \label{fig:mass_success_rate_arm}
    \end{subfigure}
    \begin{subfigure}{.32\textwidth}
      \centering
      \includegraphics[width=.95\linewidth]{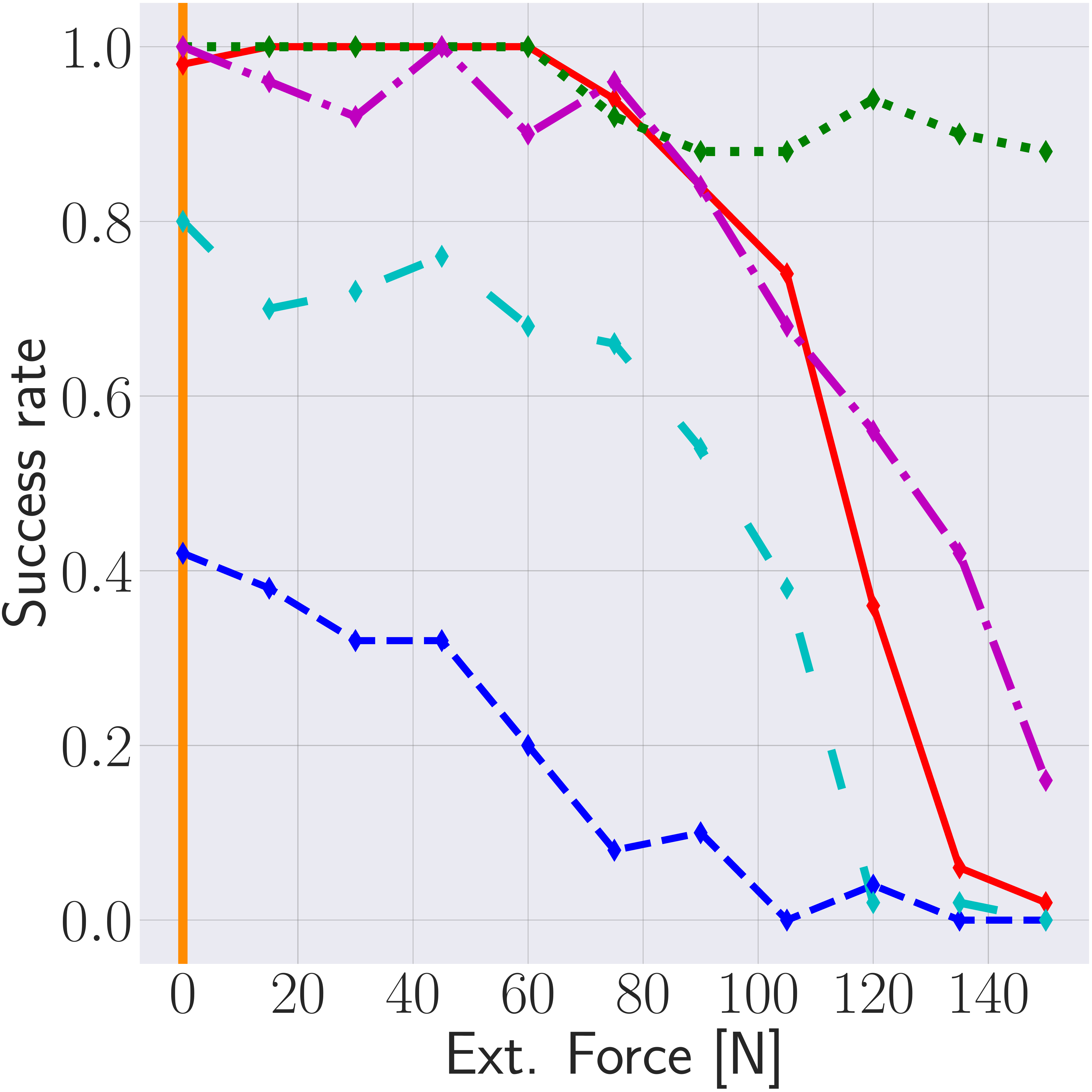}
      \caption{}
      \label{fig:extForceMagnitude_success_rate_arm}
    \end{subfigure}
    \begin{subfigure}{.32\textwidth}
      \centering
      \includegraphics[width=.95\linewidth]{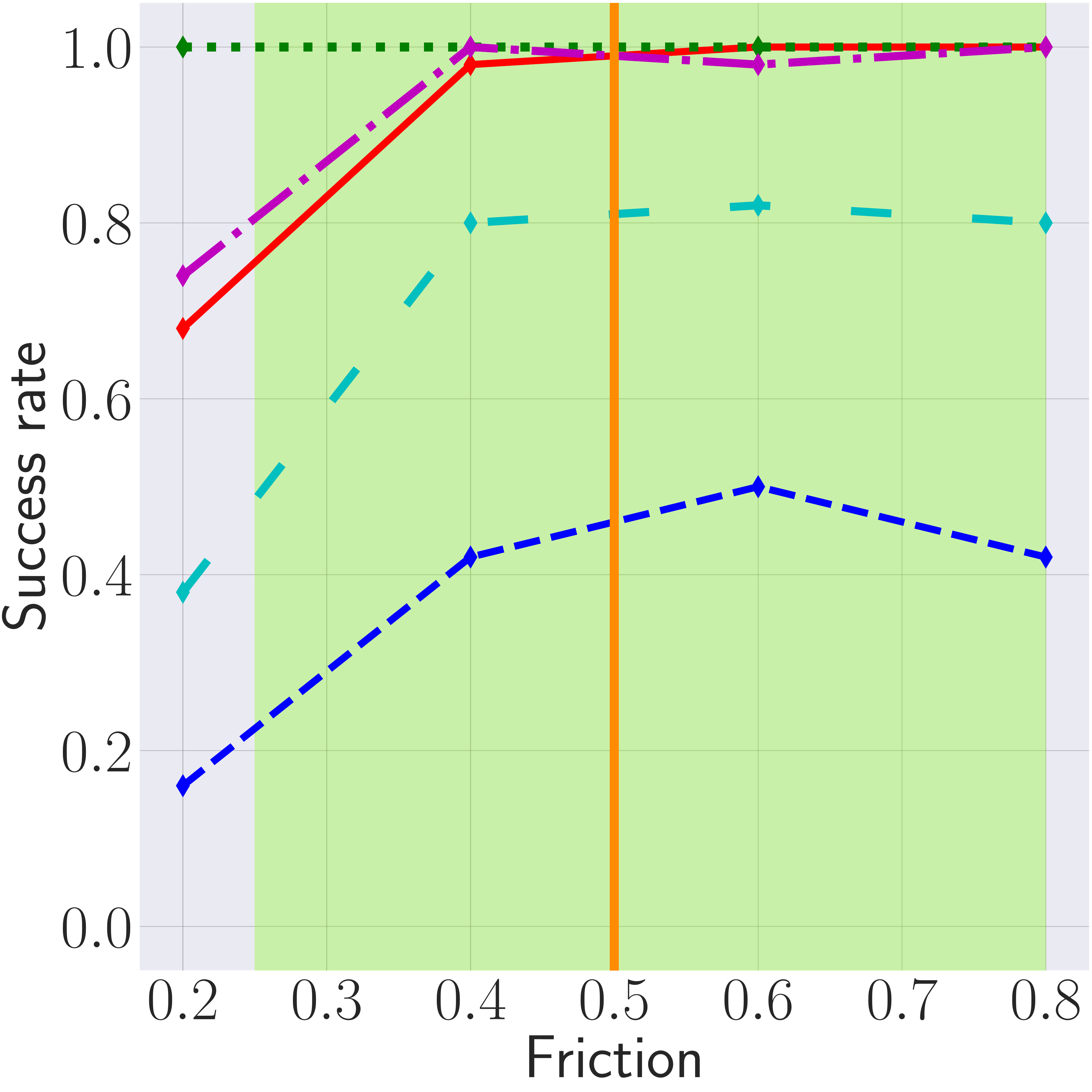}
      \caption{}
      \label{fig:friction_success_rate_arm}
    \end{subfigure}
    \\
    \begin{subfigure}{.32\textwidth}
      \centering
      \includegraphics[width=.95\linewidth]{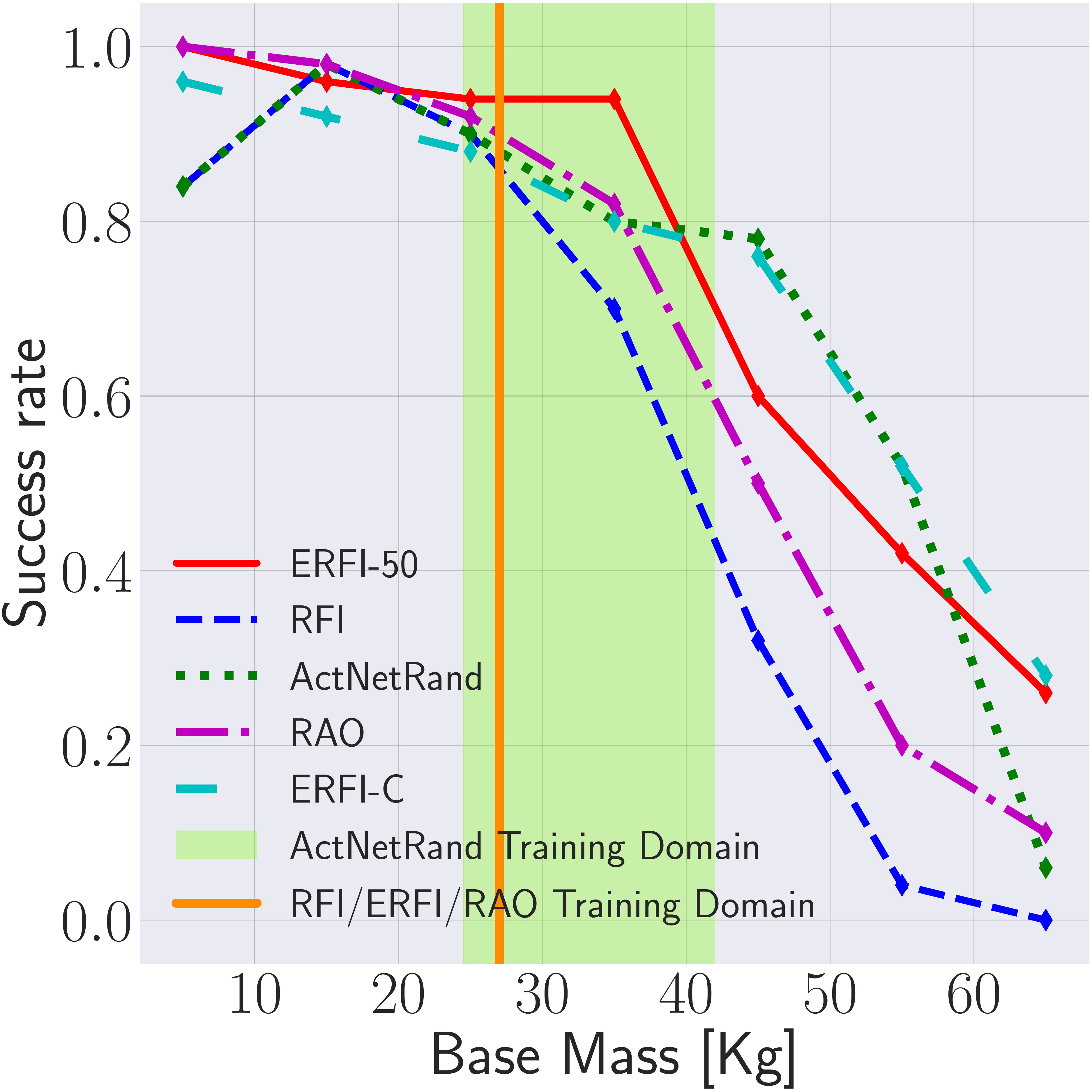}
      \caption{}
      \label{fig:mass_success_rate_rocky}
    \end{subfigure}
    \begin{subfigure}{.32\textwidth}
      \centering
      \includegraphics[width=.95\linewidth]{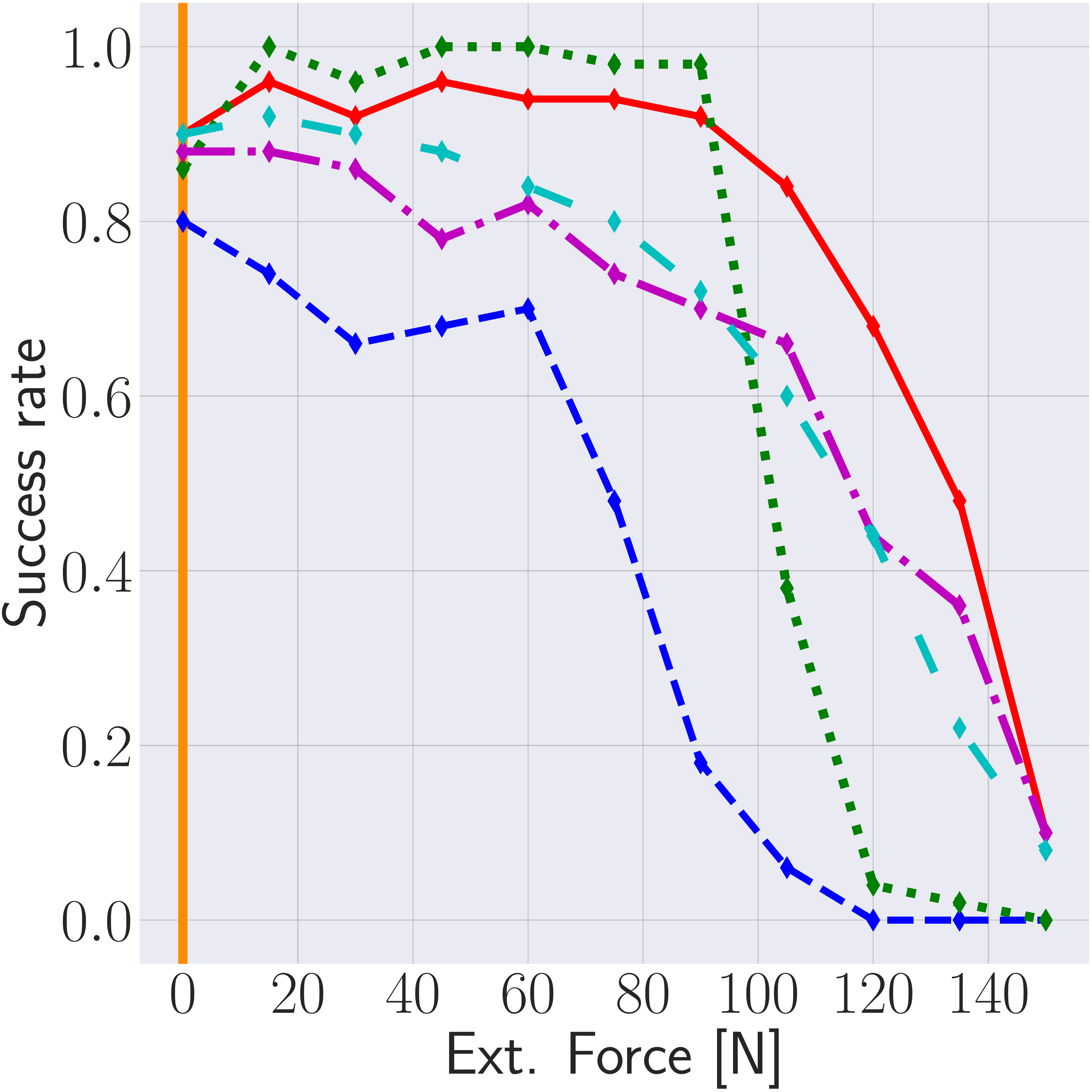}
      \caption{}
      \label{fig:extForceMagnitude_success_rate_rocky}
    \end{subfigure}
    \begin{subfigure}{.32\textwidth}
      \centering
      \includegraphics[width=.95\linewidth]{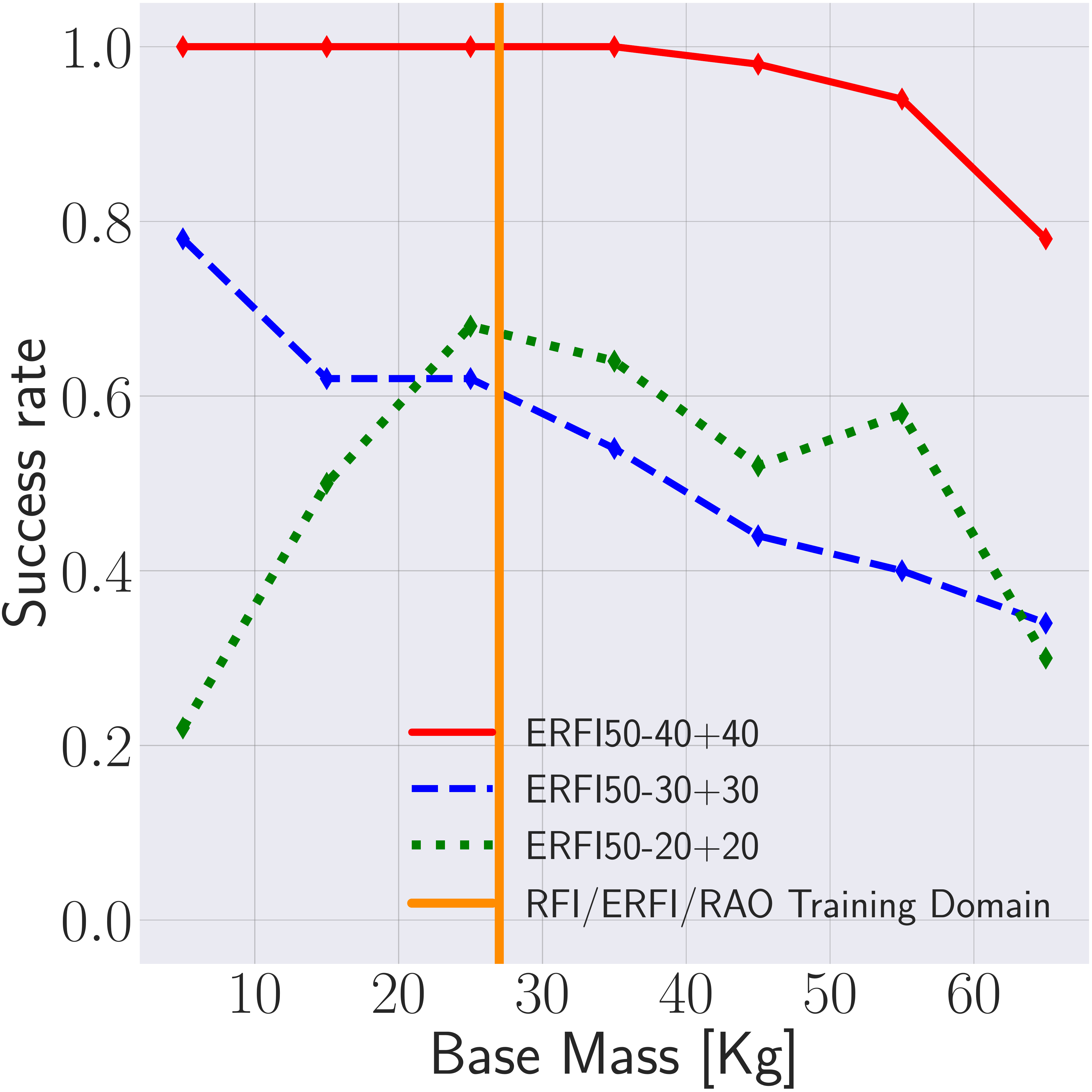}
      \caption{}
      \label{fig:ERFI_different_magnitudes}
    \end{subfigure}
    \caption{\Cref{fig:mass_success_rate_no_arm,fig:extForceMagnitude_success_rate_no_arm,fig:friction_success_rate_no_arm} show how \gls{rfi}, \gls{erfi50}, \gls{erfic}, \gls{rao} and ActNetRand resist to variations of the base mass, to external forces, or to different frictions. In \Cref{fig:mass_success_rate_arm,fig:extForceMagnitude_success_rate_arm,fig:friction_success_rate_arm} the same experiments are replicated with a Kinova manipular on top of the robot. In \Cref{fig:mass_success_rate_rocky,fig:extForceMagnitude_success_rate_rocky} we investigated the effects of the perturbations also on the rocky terrain in \Cref{fig:test_environments}. While, in \Cref{fig:ERFI_different_magnitudes} we studied how different $\tau^{lim}_{o_j}$ and $\tau^{lim}_{r_j}$ affected the robustness of the controller.}
    \label{fig:results} 
    \vspace{-0.5cm}
\end{figure*}


\section{Results} \label{sec:results}
We compared \gls{erfi50}, \gls{erfic}, and \gls{rao} against two baselines, \gls{rfi} and ActNetRand (policy trained using dynamics randomization and actuator network). The metric adopted to assess their performances is the success rate described in \Cref{sec:experimental_design}.
The first row of Figure \ref{fig:results} (\Cref{fig:mass_success_rate_no_arm,fig:extForceMagnitude_success_rate_no_arm,fig:friction_success_rate_no_arm}) shows the robustness of the different approaches to changes in the base mass, in the application of external forces, or to different friction coefficients between feet and ground; in this first batch of experiments, the arm was not included.
From these plots, it is evident that the standard \gls{rfi} is the least performing method, while still providing decent robustness especially close to the training domain.
Conversely, \gls{rao} and \gls{erfi50} are the better-performing ones (providing on average 53\% better success rate than \gls{rfi} on mass variations, \Cref{fig:mass_success_rate_no_arm}), they are often very close and sometimes better than ActNetRand, which is currently the standard approach to deploy controllers on the hardware.
Regarding \gls{erfic}, it does better than standard \gls{rfi} (on average 41\% better success rate on mass variations, \Cref{fig:mass_success_rate_no_arm}), but still not as well as ERFI-50 and \gls{rao} (on average 12\% worse success rate on mass variations, \Cref{fig:mass_success_rate_no_arm}).
The analysis presented above was repeated after mounting a fixed Kinova manipulator arm on top of the robot; the same policies, perturbation, and set of stairs were considered during the experiments.
The objective of this last study is to test the robustness of the controller in real-world scenarios never encountered during training.
The resulting performances are depicted in \Cref{fig:mass_success_rate_arm,fig:extForceMagnitude_success_rate_arm,fig:friction_success_rate_arm}, where we observe the gap between \gls{rfi} and \gls{rao}/\gls{erfi50} enlarging with a performance loss for \gls{rfi} -even in the training domain- of roughly 50\%, while \gls{rao} and \gls{erfi50} achieved roughly 62\% higher success rate than \gls{rfi} on this task, \Cref{fig:mass_success_rate_arm}.

Furthermore, we investigated the effects of $\tau^{lim}_{o_j}$ and $\tau^{lim}_{r_j}$ on the overall performances of ERFI-50, we show the outcomes of different limits on the success rate when the base mass is increased,  \Cref{fig:ERFI_different_magnitudes}.
The curves in \Cref{fig:ERFI_different_magnitudes} show that high $\tau^{lim}_{o_j}$ provides greater robustness in combination with high $\tau^{lim}_{r_j}$ (ERFI50-40+40, red line), when compared to $\tau^{lim}_{o_j} = 20 [\si{\newton\meter}]$ and $\tau^{lim}_{r_j} = 20 [\si{\newton\meter}]$.
However, for values of $\tau^{lim}_{o_j} = 30 [\si{\newton\meter}]$ and $\tau^{lim}_{r_j} = 30 [\si{\newton\meter}]$ the performance improves in one portion of the domain and it remains comparable to $\tau^{lim}_{o_j} = 20 [\si{\newton\meter}]$ and $\tau^{lim}_{r_j} = 20 [\si{\newton\meter}]$ in the remaining one.

To provide a comprehensive assessment, we present the performance on the rocky terrain in \Cref{fig:test_environments}, which complements the results obtained in the staircase environment. Notably, the survival rates in high perturbation regimes are partially reduced due to the absence of rocky terrain in the training environment. Nonetheless, the relative performances of the methods remain comparable to those observed in the staircase evaluation, \Cref{fig:mass_success_rate_rocky,fig:extForceMagnitude_success_rate_rocky}.

The robustness to further perturbations --as varying the duration of the external force, applying an external torque, varying the gravitational acceleration and shifting the knee motors' positions-- were investigated and the results are consistent with what is already shown in  \Cref{fig:results}.

\section{Conclusion}
\label{sec:conclusion}
In this work we showed that transferring policies trained in simulation to real systems is possible without defining the domain randomization's parameters and their ranges, without further system identification to measure the noise to inject in the observations and without recording any of hardware data to train an additional \gls{nn} to model the motors' dynamics.
Instead we proposed to use a blend of episodic and continuously changing random force perturbations (\gls{erfi50}), which has competitive performance compared to state of the art extensive domain randomization (ActNetRand) and which only requires tuning two parameters ($\tau^{lim}_{o_j}$ and $\tau^{lim}_{r_j}$); hence reducing the sim-to-real transfer efforts compared to previous approaches by a large margin.
We further demonstrated the validity of our approach by transferring the controllers to the hardware and showing stable locomotion with Unitree A1, uneven terrain locomotion and mounting an unmodelled manipulator on top of the robot with ANYmal~C.

\clearpage
\bibliographystyle{IEEEtran}
\bibliography{bibliography}

\end{document}